\documentclass{article}
\usepackage{arxiv}

\usepackage[utf8]{inputenc} 
\usepackage[T1]{fontenc}    
\usepackage{hyperref}       
\usepackage{url}            
\usepackage{booktabs}       
\usepackage{amsfonts}       
\usepackage{nicefrac}       
\usepackage{microtype}      

\usepackage{graphicx}

\usepackage{esvect}
\usepackage{amsmath}
\usepackage[tight,footnotesize]{subfigure}
\usepackage{appendix}
\usepackage{lscape}
\usepackage{textcomp}
\usepackage{amssymb}
\usepackage{lscape}
\usepackage[noend]{algpseudocode}
\usepackage[boxruled]{algorithm2e}
\usepackage{hhline}
\usepackage{slashbox}
\usepackage{multirow}
\usepackage{lettrine}
\usepackage{tabularx}

\title{Backpropagation-Free Learning Method for Correlated Fuzzy Neural Networks}

\author{
  Armin Salimi-Badr \\
  Faculty of Computer Science and Engineering\\
  Shahid Beheshti University\\
  Tehran, Iran \\
  \texttt{a\_salimibadr@sbu.ac.ir} \\
   \And
 Mohammad Mehdi Ebadzadeh \\
  Department of Computer Engineering\\
  Amirkabir University of Technology\\
  Tehran, Iran \\
  \texttt{ebadzadeh@aut.ac.ir} \\
}

\begin{document}
\maketitle

\begin{abstract}
In this paper, a novel stepwise learning approach based on estimating desired premise parts' outputs by solving a constrained optimization problem is proposed. This learning approach does not require backpropagating the output error to learn the premise parts' parameters. Instead, the near best output values of the rules premise parts are estimated and their parameters are changed to reduce the error between current premise parts' outputs and the estimated desired ones. Therefore, the proposed learning method avoids error backpropagation, which lead to vanishing gradient and consequently getting stuck in a local optimum. The proposed method does not need any initialization method. This learning method is utilized to train a new Takagi-Sugeno-Kang (TSK) Fuzzy Neural Network with correlated fuzzy rules including many parameters in both premise and consequent parts, avoiding getting stuck in a local optimum due to vanishing gradient. To learn the proposed network parameters, first, a constrained optimization problem is introduced and solved to estimate the desired values of premise parts' output values. Next, the error between these values and the current ones is utilized to adapt the premise parts' parameters based on the gradient-descent (GD) approach. Afterward, the error between the desired and network's outputs is used to learn consequent parts' parameters by the GD method. The proposed paradigm is successfully applied to real-world time-series prediction and regression problems. According to experimental results, its performance outperforms other methods with a more parsimonious structure.\\

\\
The full version of this preprint is accepted for publication in Neurocomputing as: \\
\textit{A. Salimi-Badr, M.M. Ebadzadeh, "A novel learning algorithm based on computing the rules' desired outputs of a TSK fuzzy neural network with non-separable fuzzy rules," Neurocomputing, vol. 470, pp. 139-153, 2022.
DOI : 10.1016/j.neucom.2021.10.103}
\end{abstract}

\section{Introduction}
\label{section1}
Fuzzy Neural Networks (FNNs) are effective hybrid techniques, able to combine interpretability, learning capability, ability to express uncertainty and knowledge of domain experts, used to deal with many problems where other non-fuzzy approaches find it difficult or impossible to tackle \cite{Salimi2020novel,Ebadzadeh2017,Ebadzadeh15,yu2020extended,ANFIS,SOFMLS,yang2019anfis,Lv2018dynamical,Feng2020fuzzy,DENFIS,Ebadzadeh09,Ganji2019,Das15}. FNNs are a kind of adaptive expert systems which are able to learn a set of interpretable rules in the form of "IF-THEN" to model a nonlinear system. It is proved that FNNs are universal approximators \cite{Ying98,Zeng00}. The functionality of FNNs as powerful tools is investigated in different applications including data mining, signal processing, system modeling, medical applications, robotics and control \cite{yu2020extended,liu2015brain,shi2015mixed,zhang2017nonlinear,liu2019adaptive,he2017adaptive,sun2018fuzzy,Ebadzadeh08,Khodabandelou2019,rubio2019anfis,AsadiEydivand2015}.

A data-driven nonlinear modeling approach uses the observations to build a nonlinear model able to describe the relation between the inputs and outputs \cite{zhou2020nonlinear}. Two well-known Fuzzy Inference Systems (FIS) considered as the backbone of FNNs for nonlinear modeling are \textit{Mamdani} \cite{mamdani1977application} and \textit{Takagi-Sugeno-Kang} (TSK) fuzzy methods \cite{Takagi85,sugeno1988structure}, which the later is more general with a larger number of parameters. A Mamdani FIS, models a nonlinear function as a point-to-point knowledge base, while a TSK one maps each point in the input space to a line in the output space. Indeed, a TSK approach considered a nonlinear manifold partially linear.

To construct an FNN able to perform data-driven modeling, two important issues must be addressed \cite{SOFMLS,yang2019anfis,Salimi2020novel,Ebadzadeh2017,Ebadzadeh15,Ebadzadeh09}: 1- determining the number of required fuzzy rules with their initial values (\textit{structure identification}), and 2- deriving appropriate values for different parameters of the model (\textit{parameter estimation}).
To overcome these issues, various approaches have been proposed in the literature. The well-known Adaptive Network-based Fuzzy Inference System (ANFIS) \cite{ANFIS} partitions uniformly each input variable into fuzzy sets which could extremely increase the number of rules more than required. Next, the \textit{Gradient Descent} (GD) optimization method is utilized to adapt rules consequent parts' parameters.

To overcome the large number of rules problem, which causes overfitting and makes rules uninterpretable, some previous methods use some clustering approaches, like Fuzzy C-Means (FCM), K-Neares Neighbors (KNN), Gustafson-Kessel (GK) or contour-aware method to initialize rules parameters \cite{Ebadzadeh15,Malek11,Teslic11,Ebadzadeh2017,Feng2020fuzzy}. Next, some local search methods, like GD, \textit{Linear Least Square Error} (LLS), or \textit{Levenberg-Marquardt} (LM), are utilized to adapt parameter values.

Another class of algorithms, uses an online clustering approach to construct fuzzy rules encountering a data stream, and utilizes local search methods generally to fine-tune consequent parts parameters \cite{Salimi2020novel,yang2019anfis,DENFIS,EFuNN,FAOSPFNN,SOFMLS,GPFNN,Malek11,AsadiEydivand2015,Salimi-Badr2017}. These approaches add new fuzzy rules encountering new instances, based on some appropriate criteria. Generally the rules premise parts parameters are considered fixed in these methods.

Moreover, some previously presented approaches utilize meta-heuristic search methods including swarm intelligence algorithms and evolutionary algorithms for structure identification and also parameter estimation \cite{SOFNNGA,Ebadzadeh09}. Additionally Some methods apply Support Vector Machines (SVM) \cite{Juang12,Ganji2019,Khodabandelou2019}, wavelet fuzzy neuron \cite{huang2017hybrid}, and Type-2 fuzzy logic \cite{ashrafi2020it2,SEIT2FNN,eT2FIS}.

Although backpropagating the output error to adapt premise parts' parameters in the form of \textit{GD} or \textit{LM} method is usual for \textit{Mamdani} type FNNs (applied generally by removing \textit{Normalization} layer \cite{Ebadzadeh15, Ebadzadeh2017}), it is not very common for \textit{TSK} type FNNs. Generally, in \textit{TSK} type FNNs, premise parts' parameters are initialized based on one of the reviewed approaches and remain fixed during learning process \cite{yang2019anfis,zhou2020nonlinear,nguyen2015gsetsk,ashrafi2020it2,Ebadzadeh09,ANFIS}. Indeed, in \textit{TSK} type FNNs, the backpropagation paradigm can easily get stuck in local optima based on the \textit{vanishing gradient} problem due to the larger number of parameters and layers. Therefore, generally \textit{LLS} or \textit{GD} methods are utilized to fine-tune just the consequent parts' parameters after initialization.



Another important issue to improve the FNN performance in nonlinear system modeling is to consider the input variables interactive and generating non-separable fuzzy rules \cite{Zadeh75, Ebadzadeh15, Ebadzadeh2017} to model their relations. Recently, new FNN models considering the relations among input variables are proposed \cite{Lemos2011,Teslic11,Ebadzadeh15,Pratama2014a,Pratama2014b,Ebadzadeh2017}.
In \cite{Lemos2011} an evolving FNN is presented which is able to consider input variables by utilizing the \textit{Mahalanobis distance} and estimating the covariance matrix of input variables. However, the extracted fuzzy rules in this approach are not in the standard interpretable format. Moreover it must calculate the inverse matrices for every training instance and every training iteration. In \cite{Pratama2014a,Pratama2014b}, evolving correlation-aware structures are proposed which are able to extract rotated fuzzy rules based on considering correlations among input variables. Although fuzzy rules in these methods are in the standard interpretable format, it is necessary to decompose covariance matrices to their eigenvalues. In \cite{Ebadzadeh15}, we have proposed another correlation-aware structure (\textit{CFNN}). In This method, the relations among input variables in each fuzzy rule are learned by applying \textit{LM} learning paradigm, backpropating the output error to the premise parts of fuzzy rules.  Although the structure of CFNN is more compact in comparison to the previous methods, its derived fuzzy rules are not in the standard format. In \cite{Ebadzadeh2017}, we have improved our previous method and proposed a novel correlation-aware Mamdani type FNN structure with intuitive initialization approach (\textit{IC-FNN}). Our initialization method in \cite{Ebadzadeh2017}tries to extract fuzzy rules similar to the surface of the covered regions of the nonlinear function by minimizing error between their contour lines. Afterwards, \textit{LM} method is utilized to fine-tune initialized parameters by backpropagating the output error. The extracted fuzzy rules are in the standard interpretable format. However, it is based on \textit{Mamdani} FIS, not \textit{TSK} one. Therefore, it can just map a manifold in input space (based on considering relations between input variables) to a point (and not a manifold) in the output space (due to the \textit{Mamdani} FIS). Furthermore, for both initialization and fine-tuning steps, many inverse matrix calculations and matrix decompositions are required.

In this paper, a new correlation-aware TSK-FNN architecture able to consider relations between interactive input variables by extracting non-separable fuzzy rules, is proposed. To overcome the mentioned problems for learning such an architecture, a novel learning algorithm, based on finding near desired outputs of the premise parts of fuzzy rules is presented. First, a constrained optimization problem is defined and solved to extract near desired outputs of the fuzzy rules antecedent parts. Next, the premise parts parameters are adapted to minimize the error between these extracted desired outputs and the real ones. Finally, the consequent parts' parameters are fine-tuned by GD to minimize the output error. By applying this approach the premise parts parameters are not considered fixed, and also the backprogation-based methods or meta-heuristic approaches are not utilized. Moreover, the algorithm does not require any initialization algorithm and starts from random variables. The main contributions of our proposed algorithm can be summarized as follows:
\begin{enumerate}
  \item A new correlation-aware architecture based on TSK FIS is proposed which is able to map a manifold in the input space by non-separable fuzzy rules to a manifold in the output space based on TSK nature;
  \item A novel two step learning algorithm for adapting premise parts parameters based on finding the desired output values of rules layer by solving a constrained optimization problem, without backpropagating the output error;
  \item The network is able to learn fuzzy rules with surfaces similar to the covered region of the target function;
  \item Complicated initialization method and complex computations including inverse matrix calculation and matrix decomposition, are not required.
\end{enumerate}

The rest of this paper is organized as follows: First, in Section \ref{section2}, the structure of the proposed architecture and the proposed learning algorithm are discussed in details. Next, section \ref{section3} reports the effectiveness of the proposed method in approximating different nonlinear functions and time-series in comparison to the other methods. Finally, conclusions are presented in Section \ref{section4}.

\section{Proposed Method}
\label{section2}
In this section, first, we present the proposed TSK correlation-aware structure. Afterward, we explain our novel learning paradigm. Finally, the structure will be interpreted as an TSK FIS and the form of its fuzzy rules is presented.

\subsection{Proposed Architecture}

The structure of the proposed FNN composed of seven distinct layers is shown in Fig. \ref{fig_1}. The antecedent part is very similar to our previously proposed model \cite{Ebadzadeh2017}, but to realize a TSK FNN, the consequent part is different. The seven layers are \textit{"input layer"}, \textit{"transformation layer"}, \textit{"fuzzy sets layer"}, \textit{"fuzzy rules layer"}, \textit{"normalization layer"}, \textit{"consequent layer"}, and finally the \textit{"output layer"}. The mathematical description of each layer is represented as follows:

\begin{figure}[!t]
\centering
\includegraphics[width = 5.5in]{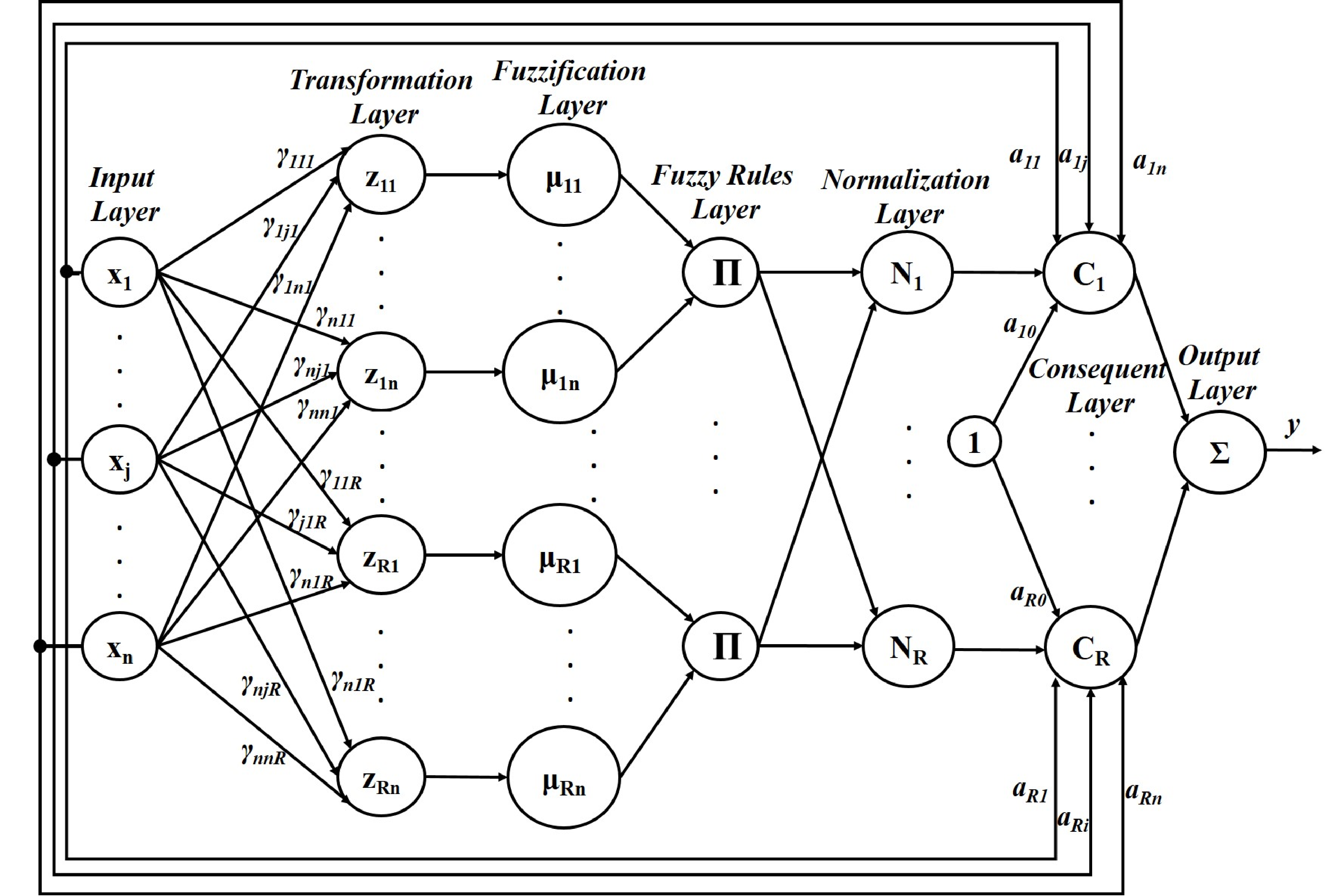}
\caption{The proposed TSK architecture.}
\label{fig_1}
\end{figure}

\begin{enumerate}
  \item \textit{The input layer:} The neurons of this layer irradiate the input variables of the network that can be expressed as the following equation:
\begin{equation}
X = [x_1,  x_2,  \cdots, x_n]^T
\end{equation}
  \item \textit{The transformation layer:} To consider the interaction between input variables in different parts of the target nonlinear function, each neuron of this layer applies a linear transformation on the input variables. As a result, the initial input space with interactive and correlated variables is mapped to different new spaces with high-level non-interactive features \cite{Ebadzadeh2017}.
  Each neuron of this layer applies a linear transformation on the input variables to map the initial interactive input variables to a new space with high-level non-interactive features. The output of this layer is calculated based on equation \eqref{eq_2}.
\begin{equation}
    Z_i = \Gamma_i^T(X-M_i)
\label{eq_2}
\end{equation}
where $M_i$ is the center of the $i^{th}$ fuzzy rule and matrix $\Gamma_i$ is the transformation matrix for $i^{th}$ fuzzy rule to transform initial input space to non-interactive feature space proper for this rule. The element in $l^{th}$ row and $m$ column of matrix $\Gamma_i$ is shown as $\gamma_{l,m,i}$.

\item \textit{The fuzzification layer}: Neurons of this layer calculates the membership degree of extracted non-interactive variables for each fuzzy rule. In order to make the shape of fuzzy sets flexible, we introduce a power value $\beta_i$ for $i^{th}$ fuzzy rule and we call this parameter "shape regulator" \cite{Ebadzadeh2017}. The membership degree for $j^{th}$ extracted variable of the $i^{th}$ fuzzy rule is computed as follows:
\begin{equation}
    \mu_{ij} =  A\left(z_{ji}, \beta_{i}\right) = e^{-\frac{1}{2}\left((z_{ji})^2\right)^{\beta_{i}}}
    \label{eq_3}
\end{equation}
where $z_{j,i}$ is the $j^{th}$ feature dimension for $i^{th}$ fuzzy rule. The main difference between this definition and our previous study \cite{Ebadzadeh2017} is that in this paper we reduce the number of parameters and limit it for each fuzzy rule ($\beta_i$) instead of for each dimension and each fuzzy rule ($\beta_{ij}$).

\item \textit{The fuzzy rules layer}: Neurons of this layer compute the membership degrees of the input instance to different fuzzy rules by applying dot-product as the \textit{"T-Norm"} operator as follows:
\begin{equation}
    \mu_i(X)=  \prod _{j=1}^{n}\mu_{ij}
    \label{eq_4}
\end{equation}
Fig. \ref{fig_2} shows different shapes of fuzzy sets and also the shape of contour-lines of different extracted fuzzy rules with different values of shape regulator $\beta$.

\begin{figure}[!t]
\centering
\includegraphics[width = 4in]{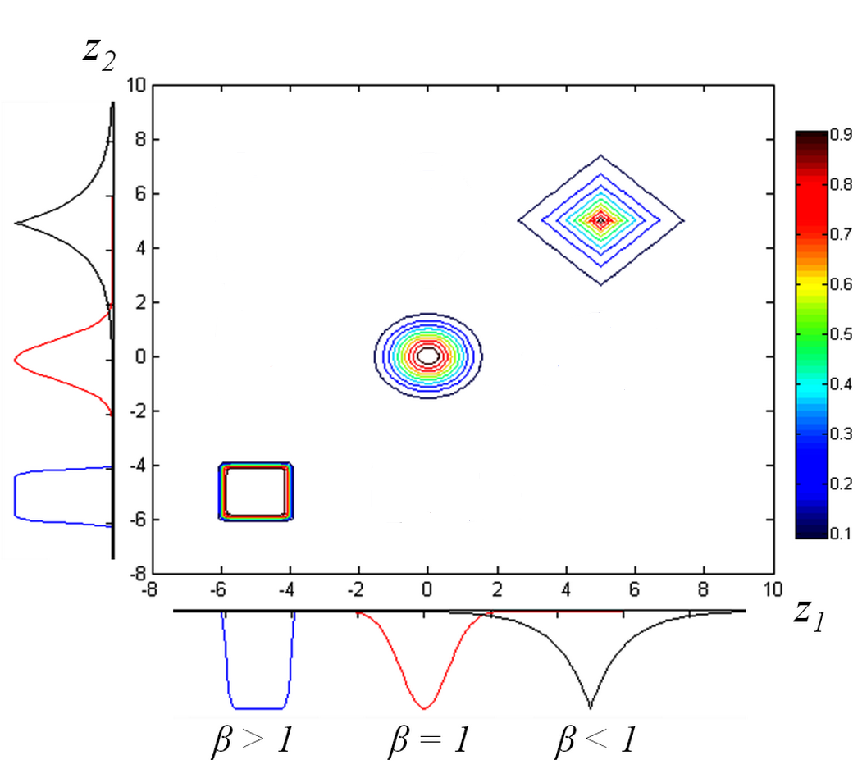}
\caption{Different shapes of fuzzy sets and fuzzy rules in the extracted feature space $Z$ with different values of the shape regulator parameter $\beta$.}
\label{fig_2}
\end{figure}

\item \textit{The normalization layer}: In this layer, the firing strength of different fuzzy rules in response to the input instance is calculated by normalizing the membership values to different fuzzy rules:
    \begin{equation}
    \phi_i(X)=  \frac{\mu_{i}(X)}{\sum _{l=1}^{R}\mu_{l}(X)}
    \label{eq_5}
\end{equation}

\item \textit{The consequent layer}: Neurons of this layer take input vector from the input layer and apply a linear function to this vector for computing the predicted output based on each fuzzy rule. Next, it weights the rule's firing strength as its final output:
    \begin{equation}
         y_i^+=  \phi_i.y_i
         \label{eq_6}
     \end{equation}
and the weight $y_i$ is computed as follows:
    \begin{equation}
         y_i=  a_{i0} + \sum_{j=1}^{n}a_{ij}.x_j
         \label{eq_7}
     \end{equation}
where $a_{ij}$ (j = 0,1, ..., n and i=1,2,..., R) is the consequent parts parameters related to the $i^{th}$ fuzzy rule and $j^{th}$ input variable.

\item \textit{The output layer}: The final output of the network is provided by the single neuron on this layer:
\begin{equation}
y = \sum_{i=1}^R y_i^+
\label{eq_8}
\end{equation}

\end{enumerate}

Similar to our previous studies \cite{Ebadzadeh15,Ebadzadeh2017} the fuzzy rules in this method cover the \textit{hill}s in the function landscape, but based on a TSK FIS model. Therefore, the $i^{th}$ fuzzy rule can be considered intuitively as follows:

\begin{align}
    \mathcal{FR}_i:&\mbox{IF } X \mbox{ belongs to the } \mbox{$i^{th}$} \mbox{ \textit{hill}} \\
    &\mbox{THEN} \mbox{ y} \mbox{ is the corresponding point on the hyperplane }  y_i(X)
\end{align}
where $i^{th}$ hill in the function landscape is $\mu_i(X)$. Therefore, it is possible to rewrite fuzzy rule $\mathcal{FR}_i$ as follows:
\begin{align}
    \mathcal{FR}_i: \mbox{IF } X \in \mu_i(X) \Rightarrow \mbox{ y} \mbox{ is} \mbox{ $y_i(X) = a_{i0} + \sum_{j=1}^{n}a_{ij}.x_j$}
\end{align}

\begin{figure}[!t]
\centering
\includegraphics[width = 6.5in]{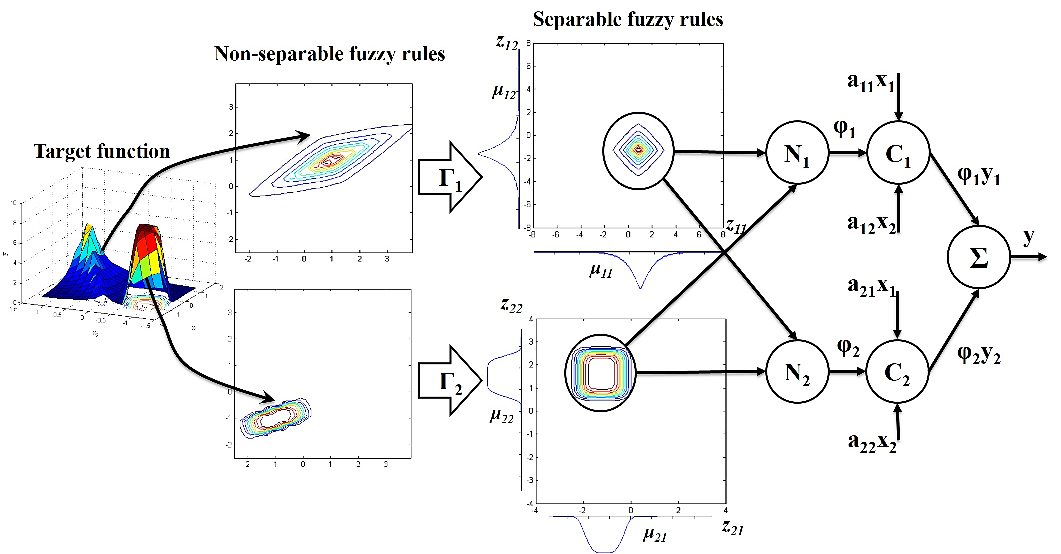}
\caption{An illustrative example to show the work-flow of the proposed network and extracting non-interactive features for different fuzzy rules from initial interactive variables.}
\label{fig_3}
\end{figure}

Fig. \ref{fig_3} shows the total work-flow of the proposed network with the process of extracting new features and non-separable fuzzy rules in these new feature spaces for interactive input variables as an illustrative example.

Suppose the proposed structure has $R$ fuzzy rules to approximate a function with $n$ input dimensions. For $i^{th}$ fuzzy rule, such a structure  has $n+1$ consequent part parameters $(a_{ij})$, $n$ parameters of the center of the fuzzy rule $(M_i)$, one \textit{regulator} parameter ($\beta_i$), and $n^2$ parameters of the transformation matrix $(\Gamma_i)$. Consequently, the number of parameters, $p$ for the proposed structure is derived as follows:
\begin{equation}
    p = R((n+1) + n + 1 + n^2) = 2R + 2Rn + Rn^2
    \label{eq_9}
\end{equation}

\subsection{Proposed Learning Method}

The proposed learning paradigm has two levels: 1- changing the premise parts' parameters based on minimizing the error between the current premise parts' outputs and desired ones by considering consequent parts' parameters fixed, and 2- changing the consequents' parts parameters to minimize the error between network's outputs and the desired ones by keeping premise parts parameters fixed. Therefore, our proposed learning method for the presented TSK correlation-aware architecture is composed of three steps:
\begin{enumerate}
  \item Finding the premise parts' desired outputs proper to minimize the network's output error with the current consequent parts' parameters;
  \item Changing premise parts parameters to minimize the error between the current rules outputs and the desired ones found in the previous step;
  \item Changing the consequent parts parameters to minimize the network's output error.
\end{enumerate}

To find the desired values of premise parts' outputs, we defined the following optimization problem ($\mathcal{P}_1$):

\begin{equation}
    \mathcal{P}_1: \left\{
    \begin{array}{ll}
              \textit{Min } & \mathcal{J}_1 = \frac{1}{2}\sum_{k=1}^{N}\sum_{i=1}^{R} (\phi_{i,k}- \psi_{i,k})^2 \\

         \textit{Subject to:} &             \\
                            & \forall k = {1,2,...,N}: \sum_{i=1}^{R} \psi_{i,k}.y_i = y_k^*
    \end{array}
    \right.
    \label{eq_10}
\end{equation}
where $\phi_{i,k}$ is the premise part's output for $i^{th}$ fuzzy rule and $k^{th}$ training instance, $\psi_{i,k}$ is the corresponding desired premise part's output which can minimize the output error due the defined constraint, and $y_k^*$ is the desired output value for the $k^{th}$ training instance. Indeed, by solving $\mathcal{P}_1$ the premise parts' output values minimizing the networks error and close as much as possible to current premise parts' outputs are extracted.

To solve $\mathcal{P}_1$, first the new augmented objective function based on the \textit{Lagrange} method is presented as follows:
\begin{equation}
    \mathcal{L} = \frac{1}{2}.\sum_{k=1}^{N}\sum_{i=1}^{R} \left(\phi_{i,k}- \psi_{i,k}\right)^2 + \sum_{k=1}^{N}\lambda_k\left(\sum_{i=1}^{R} \psi_{i,k}.y_i - y_k^*\right)
    \label{eq_11}
\end{equation}
where $\lambda_{k}$ is the \textit{Lagrange} multiplier for $k^{th}$ constraint (k=1,2,...,N) of $\mathcal{P}_1$. It is necessary to solve the following equation for each training instance (q = 1,2, ..., N) and each fuzzy rule (l = 1,2,..., R):
\begin{equation}
    \begin{array}{ll}
    \frac{\partial \mathcal{L}}{\partial \psi_{l,q}} = 0, & \Rightarrow \psi_{l,q}- \phi_{l,q} + \lambda_q.y_l = 0 \\ & \Rightarrow \psi_{l,q} = \phi_{l,q} - \lambda_q.y_l
    \end{array}
    \label{eq_12}
\end{equation}
By substituting derived $\psi_{l,q}$ into the constraint defined in $\mathcal{P}_1$ (eq. \eqref{eq_10}) we can calculate the $\lambda_{k}$:
\begin{equation}
    \begin{array}{ll}
    \sum_{i=1}^{R} \psi_{i,k}.y_i = y_k^* \\
    \Rightarrow \sum_{i=1}^{R} \left(\phi_{i,k} - \lambda_k.y_i\right).y_i = y_k^* \\
    \Rightarrow \sum_{i=1}^{R} \phi_{i,k}.y_i - \sum_{i=1}^{R}\lambda_k.y_i^2 = y_k^* \\
    \Rightarrow \lambda_k = \frac{\sum_{i=1}^{R} (\phi_{i,k}.y_i) - y_k^*}{\sum_{i=1}^{R}y_i^2}
    \end{array}
    \label{eq_13}
\end{equation}

Finally, by substituting derived $\lambda_k$ in eq. \eqref{eq_13} we have the following equation for the desired premise part's output of $l^{th}$ fuzzy rule for the $q^{th}$ training instance:
\begin{equation}
    \psi_{l,q} = \phi_{l,q} - \frac{y_l.\left(\sum_{i=1}^{R} (\phi_{i,q}.y_i) - y_q^*\right)}{\sum_{i=1}^{R}y_i^2}
    \label{eq_14}
\end{equation}

After computing the desired premise parts' outputs ($\psi$ in \eqref{eq_14}) we must change the premise parts' parameters to minimize the error between these desired values and the current ones. To reach this aim, another optimization problem $\mathcal{P}_2$ is defined as follows:
\begin{equation}
    \mathcal{P}_2:
     \textit{Min } \mathcal{J}_2 = \frac{1}{2}\sum_{k=1}^{N}\sum_{i=1}^{R} (\mu_{i,k}- \psi_{i,k})^2
    \label{eq_15}
\end{equation}
where the objective function of $\mathcal{P}_2$, $\mathcal{J}_2$, computes the total error between fuzzy rule's output ($\mu$) and the desired premise part's output ($\psi$). The decision variables of this optimization problem are premise parts parameters including $\Gamma_i$, $M_i$, and $\beta_i$. Here we utilize the \textit{GD} minimization approach and chain rule as follows:
\begin{equation}
    \begin{array}{l}
    \forall i=1,\dots, R;\,\forall l,j=1,\dots, n:\,
    \gamma_{l,j,i}^{t+1} = \gamma_{l,j,i}^{t} + \eta_{\gamma_{l,j,i}}^t.\Delta\gamma_{l,j,i}^{t} \\
    \Delta\gamma_{l,j,i} = -\frac{\partial \mathcal{J}_2}{\partial \gamma_{l,j,i}} \\
                        = -\sum_{k=1}^{N}\left(\frac{\partial \mathcal{J}_2}{\partial \mu_{i,k}}\frac{\partial \mu_{i,k}}{\partial \mu_{i,l,k}}.\frac{\partial \mu_{i,l,k}}{\partial z_{k,l,i}}.\frac{\partial z_{k,l,i}}{\partial \gamma_{l,j,i}}\right)\\
                        = -\sum_{k=1}^{N}\left((\psi_{i,k}- \mu_{i,k}).\mu_{i,k}.(\beta_i).z_{k,l,i}^{(2\beta_i-1)}.(x_{k,j}-m_{i,j}))\right)

    \end{array}
    \label{eq_16}
\end{equation}

\begin{equation}
    \begin{array}{l}
    \forall i=1,\dots, R; \, \forall j=1,\dots, n:\,
     m_{i,j}^{t+1} = m_{i,j}^{t} + \eta_{m_{i,j}}^t.\Delta m_{i,j}^{t} \\
    \Delta m_{i,j} = -\frac{\partial \mathcal{J}_2}{\partial m_{i,j}} \\
                        = -\sum_{k=1}^{N}\left(\frac{\partial \mathcal{J}_2}{\partial \mu_{i,k}}.\sum_{l=1}^{n}\left(\frac{\partial \mu_{i,k}}{\partial \mu_{i,l,k}}.\frac{\partial \mu_{i,l,k}}{\partial z_{k,l,i}}.\frac{\partial z_{k,l,i}}{\partial m_{i,j}}\right)\right)\\
                        = -\sum_{k=1}^{N}\left((\mu_{i,k}- \psi_{i,k}).\mu_{i,k}.\beta_i.\sum_{l=1}^{n}(z_{k,l,i}^{(2\beta_i-1)}.\gamma_{l,j,i})\right)

    \end{array}
    \label{eq_17}
\end{equation}

\begin{equation}
    \begin{array}{l}
    \forall i=1,\dots, R:\,
     \beta_{i}^{t+1} = \beta_{i}^{t} + \eta_{\beta_{i}}^t.\Delta \beta_{i}^{t} \\
    \Delta \beta_{i} = -\frac{\partial \mathcal{J}_2}{\partial \beta_{i}} \\
                        = -\sum_{k=1}^{N}\left(\frac{\partial \mathcal{J}_2}{\partial \mu_{i,k}}.\sum_{l=1}^{n}\left(\frac{\partial \mu_{i,k}}{\partial \mu_{i,l,k}}.\frac{\partial \mu_{i,l,k}}{\partial \beta_{i}}\right)\right)\\
                        = -\sum_{k=1}^{N}\left((\psi_{i,k}- \mu_{i,k}).\mu_{i,k}.\sum_{l=1}^{n}(ln(z_{k,l,i}).z_{k,l,i}^{2\beta_i})\right)

    \end{array}
    \label{eq_18}
\end{equation}
where in above equations, $t$ indicates the number of premise parts' parameters learning iteration, $\Delta\gamma$, $\Delta m$, and $\Delta \beta$ are the amounts of adaption for learning parameters, different $\eta$s are the learning rate multiplications which are different for each parameter, $N$ is the number of training instances, $R$ is the number of fuzzy rules, $m_{i,j}$ is the $j^{th}$ dimension of center of $i^{th}$ fuzzy rule ($M_i$),  $\gamma_{l,j,i}$ is the element in the $l^{th}$ row and $j^{th}$ column of the transformation matrix of $i^{th}$ fuzzy rule ($\Gamma_i$), and finally $z_{k.l.i}$ represents the $l^{th}$ feature resulted from mapping $k^{th}$ training input vector for $i^{th}$ fuzzy rule. These equations are applied iteratively until the changing in the mean squared error between $\psi$s and derived $\mu$s is lower than a predefined threshold $\delta_{\mu}$ indicating the convergence.

Afterward, we must adapt the consequent parts parameters ($a_{ij}$ in eq. \eqref{eq_7}). To learn these parameters, the third optimization problem is defined as follows:
\begin{equation}
    \mathcal{P}_3:
     \textit{Min } \mathcal{J}_3 = \frac{1}{2}\sum_{k=1}^{N}e_k^2=\frac{1}{2}\sum_{k=1}^{N}(\hat{y}_{k}- y_{k}^*)^2
    \label{eq_19}
\end{equation}
where the objective function of $\mathcal{P}_3$, $\mathcal{J}_3$, computes the total error between network's output ($\hat{y}$) and the desired output ($y^*$) which is defined as $e_k = \hat{y}_{k}- y_{k}^*$ for $k^{th}$ training instance. The decision variables of this optimization problem are consequent parts parameters. The GD is utilized to minimize $\mathcal{J}_3$:
\begin{equation}
    \begin{array}{l}
    \forall i=1,\dots, R; \quad \forall j=0,\dots, n: \\
     a_{i,j}^{t^{'}+1} = a_{i,j}^{t^{'}} + \eta_{a_{i,j}}^{t^{'}}.\Delta a_{i,j}^{t^{'}} \\
    \Delta a_{i,j} = -\frac{\partial \mathcal{J}_3}{\partial a_{i,j}} \\
                        = -\sum_{k=1}^{N}\left(\frac{\partial \mathcal{J}_3}{\partial \hat{y}_{k}}.\frac{\partial \hat{y}_{k}}{\partial y_{i}}.\frac{\partial y_{i}}{\partial a_{i,j}}\right)
                        = -\sum_{k=1}^{N}\left(e_k.\phi_{i,k}.x_{k,j})\right)
    \end{array}
    \label{eq_20}
\end{equation}
where $t^{'}$ indicates the number of consequent parts' parameters learning iteration, $\Delta a$ is the amount of changing for learning these parameters, and $\eta_a$ is the learning rate multiplication for consequent parts' parameters. These equations are applied iteratively until the changing in the mean squared output error become lower than a predefined threshold $\delta_e$ indicating the convergence.

Each learning epoch is defined as solving these three optimization problems once. Solving these optimization problems $\mathcal{P}_1$ to $\mathcal{P}_3$ are performed iteratively until the amount of output error becomes lower than a predefined threshold ($\epsilon$) or the number of learning epochs reaches the predefined maximum number.

To change the learning rate, the direction of current step is compared to the average amount of previous ones to decrease learning rate if these two vectors are not in accordance. If $\rho$ is a parameter, $\eta_\rho$ is the corresponding learning rate, $\Delta \rho$ is the current amount of changing, and $\Delta \rho_{avg}$ is the average amount of applied changes on $\rho$ during learning iterations, the learning rate will changed as our previous work \cite{Salimi2020novel}:
\begin{equation}
\begin{array}{l}
    \Delta \rho^{avg} = \alpha.\Delta \rho + (1-\alpha).\Delta \rho^{avg}\\
    \eta_\rho = \zeta.\eta_\rho.\lfloor(1-sgn(\Delta \rho^{avg}.\Delta \rho))/2\rfloor \\
    + \eta_\rho.\lceil(1+sgn(\Delta \rho^{avg}.\Delta \rho))/2\rceil
\end{array}
    \label{eq_21}
\end{equation}
where $\alpha$ and $\zeta$ are in $(0,1)$ and $sgn(x)$ is the signum function. According to (\ref{eq_21}), if the production of current learning step ($\Delta \rho$) and the average of previous steps ($\Delta \rho^{avg}$) is negative, the learning rate must be decreased by rate of $\beta$ to increase convergence speed of learning. Otherwise, it remains unchanged to avoid getting stuck in local optima. Our proposed learning algorithm is summarized in Algorithm \ref{alg_1}.

\begin{algorithm}[!b]
\caption{Proposed Learning Algorithm}\label{alg_1}
\text{Initialize all parameters randomly}\;
\text{Compute $\mathcal{J}_1$ as $E_1$ (eqn. \eqref{eq_10})}\;
\For{$iter \gets 1 \quad \KwTo \quad iter_{max}$}{
    \For{$i \gets 1 \quad \KwTo \quad R$}{
    \For{$k \gets 1 \quad \KwTo \quad N$}{
     \text{Update $\psi_{i,k}$ (eqn. \eqref{eq_14})}\;
    }
    }
    \text{Set all $\eta_\rho$s in eqns. \eqref{eq_16} \KwTo \eqref{eq_18} $\eta_0$}\;
    \text{Compute $\mathcal{J}_2$ as $E_2$ (eqn. \eqref{eq_15})}\;
    \text{Set all $\Delta_\rho^{avg}$s 0}\;
    \For{$t \gets 1 \quad \KwTo \quad t_{max}$}{
          \For{$i \gets 1 \quad \KwTo \quad R$}{
           \For{$j \gets 1 \quad \KwTo \quad n$}{
           \For{$l \gets 1 \quad \KwTo \quad n$}{
            \text{Compute $\Delta \gamma_{l,j,i}$ eqn. \eqref{eq_16}}\;
           }
           \text{Compute $\Delta m_{i,j}$ eqn. \eqref{eq_17}}\;
           }
           \text{Compute $\Delta \beta_{i}$ eqn. \eqref{eq_18}}\;
           }
    \text{Update $\gamma$s, $m$s and $\beta$s (eqns. \eqref{eq_16} \KwTo \eqref{eq_18})}\;
    \text{Update $\Delta\gamma^{avg}$s, $\Delta m^{avg}$s and $\Delta \beta^{avg}$s (eqn. \eqref{eq_21})}\;
    \text{Update $\eta_\gamma$s, $\eta_m$s and $\eta_\beta$s (eqn. \eqref{eq_21})}\;
    \text{Compute new $\mathcal{J}_2$ as $E_2^{new}$ (eqn. \eqref{eq_15})}\;
    \If{$|E_2-E_2^{new}|<\delta_{\mu}$}{
    \text{Break}\;
    }
     $E_2 \gets E_2^{new}$\;

    }
    \text{Update all $\Delta_a^{avg}$s to 0}\;
    \text{Compute $\mathcal{J}_3$ as $E_3$ and all $e_k$s (eqn. \eqref{eq_19})}\;
    \For{$t^{'} \gets 1 \quad \KwTo \quad t^{'}_{max}$}{
         \For{$i \gets 1 \quad \KwTo \quad R$}{
           \For{$j \gets 0 \quad \KwTo \quad n$}{
           \text{Compute $\Delta a_{i,j}$ eqn. \eqref{eq_20}}\;
           }
           }
           \text{Update $a$s (eqn. \eqref{eq_20})}\;
           \text{Update $\Delta a^{avg}$s and $\eta_a$s (eqn. \eqref{eq_21})}\;
           \text{Compute new $\mathcal{J}_3$ as $E_3^{new}$ and all $e_k$s (eqn. \eqref{eq_19})}\;
           \If{$|E_3-E_3^{new}|<\delta_e$}{
        \text{Break}\;
        }
        $E_3 \gets E_3^{new}$\;

        }
        \text{Compute new $\mathcal{J}_1$ as $E_1^{new}$ (eqn. \eqref{eq_10})}\;
        \If{$|E_1-E_1^{new}|<\epsilon$}{
        \text{Break}\;
        }
        $E_1 \gets E_1^{new}$\;

        }
\end{algorithm}
\section{Experiments}
\label{section3}
In this section the performance and architecture size of the proposed method (indicated as \textit{TSK-ICFNN}(\underline{T}akagi-\underline{S}ugeno-\underline{K}ang \underline{FNN} with \underline{I}nterpretable Fuzzy Rules)) is compared to those of the other methods in the literature. We investigate the performance of our method in five real-world problems and report the results in two sections: 1- Regression problems, and 2- Time-series prediction. To show that our learning approach is effective in comparison to the output backpropagation, we train the proposed structure with backpropagation algorithm (backpropagating the output error in a batch learning manner with parameter values equal to those of the proposed method) and consider it in comparisons with the name \textit{BPFNN}.

In Time-Series prediction tasks we assume that the learning objective is a one-step-ahead prediction task \cite{ANCFIS,Ebadzadeh15,Ebadzadeh2017}. In order to measure the precision and compare with the other methods, \textit{Root Mean Squared Error} (RMSE) (\cite{SOFMLS,Das15,Ebadzadeh15,Rubio15,Mansouri16,Ebadzadeh2017}) is used which is defined as follows:
\begin{align}
    RMSE & = \sqrt{\frac{1}{N}\sum_{k=1}^N\left(y_k^*-\hat{y}_k\right)^2}
\end{align}
In all experiments the reported RMSE is the average value of 10 times running the algorithm. The parameters of learning method are set: $\eta_0 = 10^{-3}$, $\delta_{\mu} = \delta_{e} = 10^{-3}$, $\alpha = 0.7$, $\beta = 0.9$. For \textit{Abalone} and \textit{California Housing} benchmarks, considering its larger number of instances, we set $\eta_0 = 10^{-4}$.

\subsection{Regression Problems}
Three real-world regression problems including a small-scale (\textit{"Auto-MPG"}), a medium-scale (\textit{"Abalone"}) and a large-scale (\textit{"California Housing"}) benchmark problems are used in these experiments. In the first benchmark, "Auto-MPG", the network must predict the fuel consumption (miles per gallon) of different models of cars based on their characteristics \cite{FAOSPFNN}. The aim of "Abalone" benchmark problem is to predict the age of \textit{Abalone} (some kind of sea snails) based on physical measurements \cite{FAOSPFNN}. Finally, "California Housing" benchmark composed of over 20000 observations for predicting the price
of houses in California based on their different specifications \cite{Ebadzadeh15}. Moreover, to compare the performance of the proposed method to larger number of previous studies, we follow two different instructions for splitting the data samples to train and test datasets (case1 and case2 in Table \ref{table_info}). The test RMSE and number of neurons of different methods for \textit{case1} are presented in Table \ref{table_results1} and for \textit{case2} in Table \ref{table_results2}.

\begin{table}[]
    \centering
    \caption[c]{\\Datasets details}
    \small
    \resizebox{\textwidth}{!}{
    \begin{tabular}{|c|c|c|c|c|c|c|}
    \hline
        Problem & Type &no. attributes & no. instances & no. training samples & no. test samples & references \\
    \hline
        Auto MPG  (case1) & Small-size regression & 6 & 392 & 320 & 72 & \cite{FAOSPFNN}\\
    \hline
        Auto MPG  (case2) & Small-size regression & 6 & 392 & 196 & 196 & \cite{Feng2020fuzzy}\\
    \hline
        Abalone & Medium-scale regression  & 8 & 4177 & 3000 & 1177 & \cite{FAOSPFNN,Das15,Ebadzadeh2017,Feng2020fuzzy}\\
    \hline
        California housing (case1) & Large-scale regression & 8 & 20640 & 8000 & 12640 & \cite{GGAPRBF,Ebadzadeh15,Ebadzadeh2017} \\
    \hline
        California housing (case2) & Large-scale regression & 8 & 20640 & 10320 & 10320 & \cite{Feng2020fuzzy} \\
    \hline
        Google stock tracking  & Time-series prediction & 3 & 1534 & 1534 & 1534 & \cite{Das15,Ebadzadeh2017} \\
    \hline
        Sydney stock tracking  & Time-series prediction & 4 & 8534 & 1260 & 7274 & \cite{ashrafi2020it2,Pratama2014a} \\
    \hline
    \end{tabular}
    }
    \label{table_info}
\end{table}
\begin{table}[t]
    \caption[c]{\\Comparison performance of the proposed method with previous methods in real-world regression problems}
    \centering
    \resizebox{\textwidth}{!}{
    \begin{tabular}{|l||c|c||c|c||c|c|}
        \hline
       \multirow{2}{*}{\backslashbox{Method}{Benchmark}} & \multicolumn{2}{c||}{Auto-MPG}& \multicolumn{2}{c||}{Abalone} & \multicolumn{2}{c|}{California Housing} \\
        \cline{2-7}
       & no. neurons & RMSE & no. neurons & RMSE & no. neurons & RMSE\\
        \hline
        TSK-ICFNN & \textbf{2}  &\underline{\textbf{ 0.0697}} & \textbf{3} & \underline{\textbf{0.0719}} & \textbf{2}  & \textbf{0.0622} \\
        \hline
        BPFNN & 2  & 0.0856 & 3  & 0.1008 & 2 & 0.1122 \\
        \hline
        IC-FNN \cite{Ebadzadeh2017} & 2  & 0.1232 &3  & 0.0774 & 2 & \underline{\textbf{0.06126}} \\
        \hline
        CFNN \cite{Ebadzadeh15} & 2 & 0.0778 & 3  & 0.0794 & 2 &  0.06874 \\
        \hline
        RAN \cite{RAN} &4.44  & 0.3080& 346  & 0.0978 & 3552  & 0.15305  \\
        \hline
        RANEKF \cite{RANEKF} &17.2  & 0.1387& 409  &  0.0794 & 200  & 0.14952  \\
        \hline
        MRAN \cite{MRAN} &4.46  & 0.1376& 88  & 0.0837 & 64  & 0.15859  \\
        \hline
        GGAP-RBF \cite{GGAPRBF} &3.12  & 0.1404& 24  &  0.0966 & 18  & 0.13861 \\
        \hline
        OS-ELM \cite{OSELM}&25  & 0.0759& 25  & 0.0770  & NA$^\dag$  & NA$^\dag$ \\
        \hline
        FAOS-PFNN \cite{FAOSPFNN} &2.9  & 0.0775& 5  & 0.0807  & NA$^\dag$  & NA$^\dag$ \\
        \hline
        Fuzzy BLS \cite{Feng2020fuzzy} &NA$^\dag$  & NA$^\dag$& 61  & 0.0745  & NA$^\dag$  & NA$^\dag$ \\
        \hline
        \multicolumn{7}{c}{$^\dag$NA: Data is not available based on the reported results}\\
    \end{tabular}
    }
    \label{table_results1}
\end{table}

According to the reported results in both Tables \ref{table_results1} and \ref{table_results2} it is shown that both performance and architecture size of our method are better than other approaches (size is not reported in Table \ref{table_results2} because it is not reported in the previous articles for methods compared in this table). In the small-scale Auto-MPG benchmark, the performance of our new method is the best, but far better than our previous correlation-aware IC-FNN \cite{Ebadzadeh2017}. We can conclude that in such a small-scale problem the number of training instances are not sufficient for the proposed initialization method of IC-FNN to extract \textit{contour-sets} of the function by its fuzzy rules. Therefore, considering the better or close performance of our new approach in other benchmark problems, we can propose that our new approach is more general than IC-FNN. Moreover, by comparing the results to another correlation-aware structure, CFNN \cite{Ebadzadeh15}, it is shown that the TSK architecture and providing fuzzy sets with adaptive shapes (considered in our new proposed method) are more efficient in different problems. Furthermore, by comparing the performance of the proposed method with performance of BPFNN, it is concluded that the proposed learning approach is more effective than backpropropagating the output error for learning premise parts' parameters in a TSK architecture. Finally, the importance of considering the correlation of input variables in fuzzy rules could be investigated by comparing the performance of the proposed method and other similar correlation-aware approaches (IC-FNN, CFNN, and BPFNN) with other non-correlation aware methods.

\subsection{Time-Series Prediction}
In this part, the performance of the proposed method is checked on predicting two real-world time series: 1- Google stock price tracking, and 2- Sydney stock price tracking. Data is collected from Yahoo finance\footnote{http://finance.yahoo.com/}. For google stock, 1534 samples of daily stock prices are collected over a period of six years from 19-August-2004 to 21-September-2010, all data is used for training and test phases. The network must predict the day-ahead price based on previous three-day data \cite{Das15,Ebadzadeh2017}. For Sydney stock tracking, 8540 daily stock prices are collected from 3rd January 1985 to 14th November 2018, samples related to the first five years are used as training data and other 28 years for the test \cite{ashrafi2020it2}. The network must predict the day-ahead price based on previous four-day data \cite{ashrafi2020it2}.

\begin{figure}[t]
\centering
\begin{tabular}{cc}
    \subfigure[][]{\includegraphics[width = 3in]{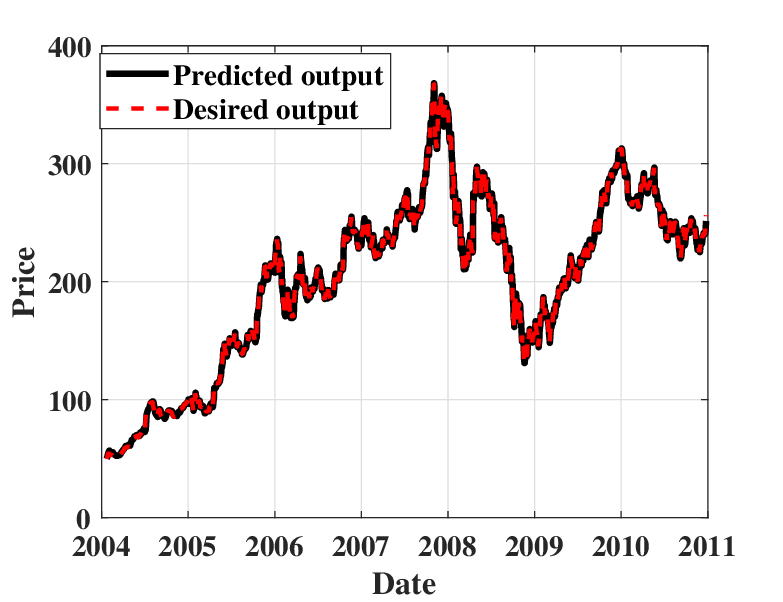}} &
    \subfigure[][]{\includegraphics[width = 3in]{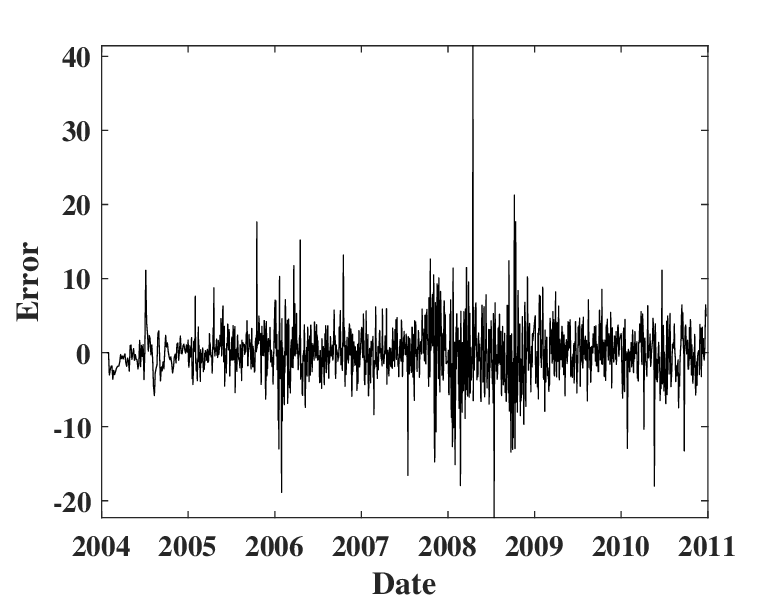}}
\end{tabular}
\caption{Comparison between actual and predicted outputs for Google stock price tracking task (a) and the amount of error (b).}
\label{fig_4}
\end{figure}

\begin{figure}[!t]
\centering
\begin{tabular}{cc}
    \subfigure[][]{\includegraphics[width = 3in]{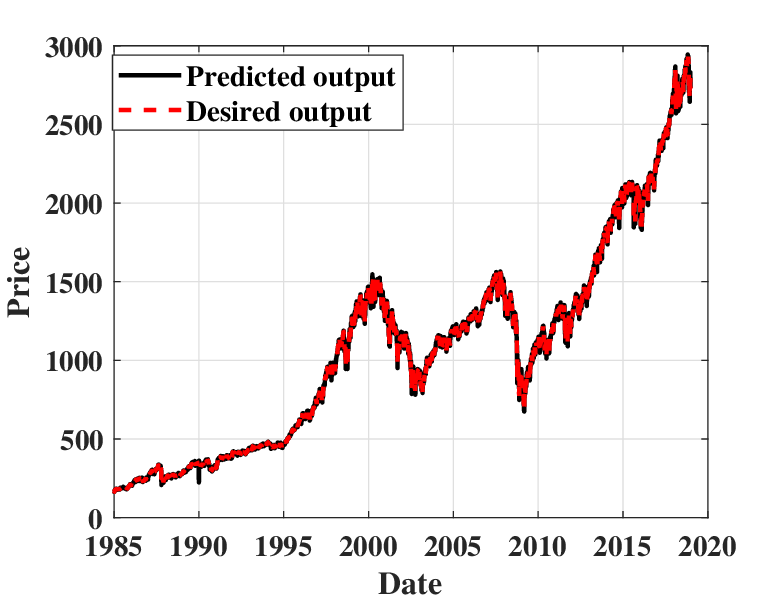}} &
    \subfigure[][]{\includegraphics[width = 3in]{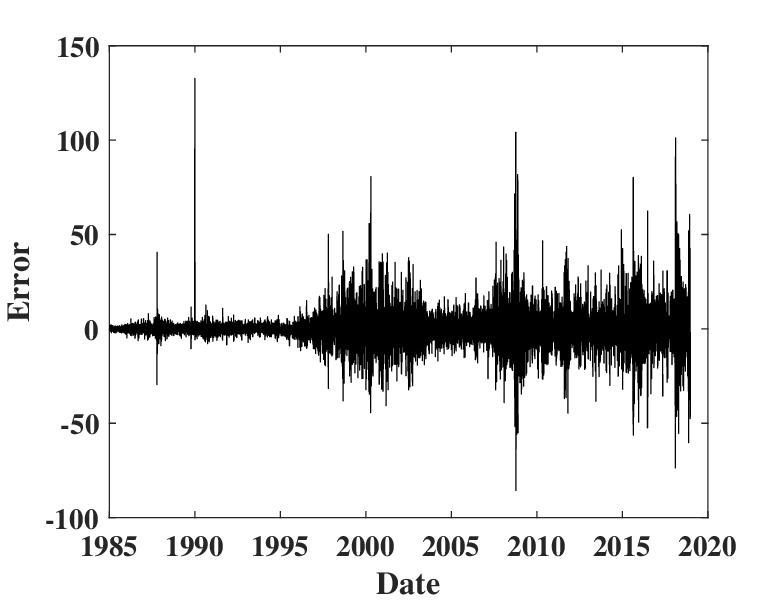}}
\end{tabular}
\caption{Comparison between actual and predicted outputs for 33 years daily Sydney stock price tracking task (a) and the amount of error (b).}
\label{fig_5}
\end{figure}

Tables \ref{table_results3} and \ref{table_results4} compare the performance and architecture size of the proposed method with other ones for time-series prediction problems. It is shown that the proposed network has the best performance with the most compact structure. Moreover, Fig. \ref{fig_4} and \ref{fig_5} compare the network's output with the desired time-series. It is shown that the network's output for each time-series is very close to desired one.

\begin{table}[t]
    \caption[c]{\\Comparison performance (test RMSE) of the proposed method with previous methods in real-world regression problems (case2)}
    \centering
    \resizebox{\textwidth}{!}{
    \begin{tabular}{|l||c|c|c|c|c|c|}
        \hline
       \multirow{2}{*}{\backslashbox{Benchmark}{Method}} & TSK-ICFNN & ANFIS \cite{ANFIS}& Fuzzy BLS \cite{Feng2020fuzzy}& OS-Fuzzy-ELM \cite{OSFELM} & Simpl\_eTS \cite{simpl_eTS} & BPFNN \\&&&&&&\\
        \hline
         Auto-MPG &  \underline{\textbf{0.0713}}  & 0.0803 & 0.0739 &0.0765 & 0.0806 &  0.0906\\
        \hline
        California Housing &  \underline{\textbf{0.0742}} &  0.1380 &  0.1283 & 0.1320 & 0.1616 &  0.0752 \\
        \hline
    \end{tabular}
    }
    \label{table_results2}
\end{table}

\begin{table}[t]
    \caption[c]{\\Comparison performance of the proposed method with previous methods in Google stock price tracking}
    \centering
    \resizebox{\textwidth}{!}{
    \begin{tabular}{|l||c|c|c|c|c|c|c|c|c|}
        \hline
       \multirow{2}{*}{\backslashbox{Measure}{Method}} & TSK-ICFNN & IC-FNN \cite{Ebadzadeh2017}& CFNN \cite{Ebadzadeh15}& BPFNN & PCA-FNN \cite{Ebadzadeh2017} & McIT2FIS-UM \cite{Das15} & McIT2FIS-US \cite{Das15} & eT2FIS \cite{eT2FIS} \\&&&&&&&&\\
        \hline
         RMSE & \underline{\textbf{3.70}} & 6.38 & 15.74& 9.22 & 65.01 & 11.1 & 9.9 & 18.46 \\
        \hline
        no. neurons & 3 & 3& 3& 3 & 3 & 7 & 11 & 34 \\
        \hline
    \end{tabular}
    }
    \label{table_results3}
\end{table}

\begin{table}[!t]
    \caption[c]{\\Comparison performance of the proposed method with previous methods in Sydney (S\&P-500) stock price tracking}
    \centering
    \resizebox{\textwidth}{!}{
    \begin{tabular}{|l||c|c|c|c|c|c|c|c|c|c|}
        \hline
       \multirow{2}{*}{\backslashbox{Measure}{Method}} & TSK-ICFNN & IC-FNN \cite{Ebadzadeh2017}& CFNN \cite{Ebadzadeh15}& BPFNN & PCA-FNN \cite{Ebadzadeh2017} & IT2GSETSK \cite{ashrafi2020it2} & GSETSK \cite{nguyen2015gsetsk} & DENFIS \cite{DENFIS} & EFuNN \cite{EFuNN} \\&&&&&&&&&\\
        \hline
         RMSE & \underline{\textbf{10.96}} & 34.37& 11.23 & 42.36 & 635.49 & 13.30 & 12.30 & 14.70 & 217.10 \\
        \hline
        no. neurons & 2 & 2 & 2 & 2 & 3 & 5 & 2 & 46&5 \\
        \hline
    \end{tabular}
    }
    \label{table_results4}
\end{table}

Based on the results reported in Table \ref{table_results3}, by comparing the performance of the BPFNN (our proposed architecture learned by BP algorithm) with other methods, the efficiency of the proposed TSK correlation-aware structure could be observed. Moreover, it is shown that the performance of the proposed method (TSK-ICFNN) is significantly better than the other methods which reveals the impression of the proposed learning method.

According to the results presented in Table \ref{table_results4}, the performance of the proposed method is better than the others. Here, by comparing the performance of the TSK-ICFNN and BPFNN, the efficiency of the proposed learning method is evidently revealed. Furthermore, the performance of our new method is far better than our previous method, IC-FNN \cite{Ebadzadeh2017}, and we can conclude that the sophisticated initialization approach of IC-FNN to extract initial fuzzy rules with contours similar to the covered region contours is not effective in this benchmark. Moreover, although the performance of CFNN is very close to the performance of our new approach, our new method does not utilize complex calculations including matrix decomposition and inverse matrix which are necessary calculations used in CFNN.

Moreover, in both benchmark studies, investigating the performance of \textit{PCA-FNN} \cite{Ebadzadeh2017}, shows that considering the correlation among input variables in whole input space is not sufficient for improving the performance of function approximation, and we must consider the relations among input variables by each fuzzy rule covering each local region of the target function, as performed in our proposed method.

\section{Conclusions}
\label{section4}
In this paper, a TSK Correlation-aware fuzzy neural network is presented. The proposed model can consider the interaction of input variables and adapts the shape of fuzzy sets and fuzzy rules. Consequently, it can extract non-separable fuzzy rules which are appropriate for interactive input variables.

One important difference between the proposed network and other related correlation-aware structures, like CFNN \cite{Ebadzadeh15} and IC-FNN \cite{Ebadzadeh2017}, is its TSK architecture. Based on the TSK nature, the proposed network is able to map a manifold in the input space to a manifold (here a hyperplane) in the output space. This ability is more general than the capability of other correlation-aware \textit{Mamdani} structures which are able to map a manifold in input space to a point in the output space.

Moreover, we present a novel learning approach to learn our proposed network's parameters efficiently. Contrary to learning paradigms used in related previous methods, our proposed learning method does not backpropagate the network error for training the premise parts' parameters. Instead, our proposed learning algorithm first searches for desired premise parts outputs. Next, it changes the network parameters to minimize the error between the current premise parts outputs and the found desired ones.

Another important advantage of our proposed learning algorithm is that it does not require any complex numerical calculations including inverse matrix computation and matrix decomposition which are parts of learning algorithms used in other related methods like CFNN \cite{Ebadzadeh15} and IC-FNN \cite{Ebadzadeh2017}. Furthermore, our presented method starts learning from random initial variables and no complex or sophisticated initialization paradigm like initialization method proposed in the recently proposed related study (IC-FNN) is used.

To show the effectiveness of our proposed method its performance is compared with the existing methods in different real world problems including different regression problems and time-series predictions. According to the experimental results, our proposed learning algorithm is more efficient in approximating nonlinear functions. Moreover, the advantage of our proposed learning algorithm over standard backpropagation algorithm is investigated to show its efficiency. Furthermore, its performance is compared with other correlation-aware methods including IC-FNN and CFNN. Its performance in all benchmark problems is better than performance of other approaches with the most parsimonious structure. In some benchmark studies like the \textit{Sydney stock price tracking} and \textit{Auto-MPG} regression problem, it is shown that the performance of our proposed method is far better than IC-FNN (as the most related method) which tries to model contour lines of the target function in its rules initialization method.

Hence in our proposed model it is assumed that all training samples are available during learning step, there are some real applications, such as sequence learning that require online-learning. Therefore, developing an online version of the proposed method able to encounter dynamic environments, is proposed as one of the future work of this study.

Moreover, as another future trend it is proposed to expand the proposed learning approach for Deep FNN structures include stacks of fuzzy rules as another future work of this paper. Since the learning method does not utilize the output error backpropagation and avoid getting stuck in local optima due to the vanishing gradient problem it could be beneficial for Deep FNN structures.

\bibliographystyle{plain}
\bibliography{ref}

\end{document}